\documentclass[runningheads,a4paper]{llncs}

\usepackage{subfigure}
\usepackage{amsmath}
\usepackage{diagbox}
\usepackage{float}
\usepackage{bm}

\usepackage{multirow}
\usepackage{amssymb}
\usepackage{graphicx}
\usepackage{tabu}

\usepackage{url}
\urldef{\mailsa}\path|{alfred.hofmann, brigitte.apfel, ursula.barth, christine.guenther,|
\urldef{\mailsb}\path|ingrid.haas, frank.holzwarth, anna.kramer, leonie.kunz, nicole.sator,|
\urldef{\mailsc}\path|erika.siebert-cole, peter.strasser, lncs}@springer.com|
\newcommand{\keywords}[1]{\par\addvspace\baselineskip
\noindent\keywordname\enspace\ignorespaces#1}

\begin{document}

\mainmatter  

\title{CNN Based Hashing for Image Retrieval}

\titlerunning{CNN Based Hashing for Image Retrieval}

%
%
\author{Jinma Guo%
\and Jianmin Li}
%

\institute{Tsinghua University, Beijing, China} 


%
%

\toctitle{Lecture Notes in Computer Science}
\tocauthor{Authors' Instructions}
\maketitle

\begin{abstract}
Along with data on the web increasing dramatically, hashing is becoming more and more popular as a method of approximate nearest neighbor search. Previous supervised hashing methods utilized similarity/dissimilarity matrix to get semantic information. But the matrix is not easy to construct for a new dataset. Rather than to reconstruct the matrix, we proposed a straightforward CNN-based hashing method, i.e. binarilizing the activations of a fully connected layer with threshold $0$ and taking the binary result as hash codes. This method achieved the best performance on CIFAR-10 and was comparable with the state-of-the-art on MNIST. And our experiments on CIFAR-10 suggested that the signs of activations may carry more information than the relative values of activations between samples, and that the co-adaption between feature extractor and hash functions is important for hashing.
\end{abstract}

\section{Introduction}
With the fast development of Social Network Service (SNS) and much easier access to camera devices, the amount of visual data on the web such as images and videos is increasing rapidly. The extremely tremendous dataset has exerted great pressure on the computation and storage efficiency of systems. Visual information retrieval is becoming more and more challenging. Recently, the need of efficient retrieval for images or videos has attracted extensive attentions from academia and industry \cite{IR}.

Exhaustive nearest neighbor search is intractable. Methods that can return satisfactory results within logarithmic ($O(log(n)$)or even constant ($O(1)$) time are more attractive. Approximate nearest neighbor search (ANN) can achieve this by organizing data with structures that keep distance metric as well as reduce the search space of each step. Tree-based approaches, including KD tree \cite{MultiTree}\cite{BestMatches}\cite{KD-trees}, ball tree \cite{Eff} and metric tree \cite{MetricTree}, can organize data with efficient structures which can boost the retrieval speed to $O(log(n))$. However, when it comes to high-dimensional data space where the samples are exponentially sparse along each dimension, the performances of these methods will be degraded even to be linear with the dataset capacity \cite{NNHD}. In addition, the memory consumed by these tree-based structures is bigger than that of the original representations in many cases \cite{MT}.

On the other hand, hashing-based methods \cite{SSHDH}\cite{KLSH}\cite{KSH}\cite{BRE}\cite{CNNH} with lookup tables consume only constant time on a new query. The compact codes of hashing can also bring down the demand of disk to store, and the bitwise operations when a query is compared with dataset make them competent even in the case of exhaustive ranking. 

Previous hashing methods take low-level features as input and use shallow models to generate the hash codes, based on the prerequisite that the visual similarity is somehow embedded in the low-level feature space. But the well-known semantic gap suggests that the distance in perception is not maintained by the low-level features. Although CNNH and CNNH+ \cite{CNNH} gain a great performance boost via leveraging deep models to learn representation and hash codes, they break the learning process into two separate stages and thus may reduce the co-adaptiveness which can be of great importance for a high performance of CNN \cite{transferable}.

In the CNNH/CNNH+ model, the approximate hash codes for training data are generated from similarity matrix by coordinate descent algorithm. Therefore we can assume the bits are not carrying concrete meanings. And the layer of CNNH/CNNH+ model that generates the binary codes is fully connected to the previous layer. According to \cite{CNNdeepFlaw}, it is the space extended by activations of a certain layer, rather than the the individual units, that contains the semantic information. Thus the model can be considered as a Siamese structure cut at the final inner-product layer into two separate stage, so the hash codes of CNNH/CNNH+ can be seen as a middle layer's outputs. And it has been shown that the deep features learned by CNN are expressive to serve as good descriptors for image retrieval \cite{NeuralCodes}. So we propose that the activations of a certain layer in CNN may have the potential for hash.

We hypothesized that the signs of deep features learned in classification tasks can represent the existence of some abstract contents. Our model were trained to minimize the classification errors just like a traditional CNN and then a  certain layer's outputs were binarilized according to their signs to perform as hash codes. Thus it was an end-to-end system where feature extraction and hashing were combined.

The specific contributions of our work are as follows: as we know, we are the first to use CNN alone as a method to learn hashing with labeled data and we got the best performance on CIFAR-10 (8\% to 16\% improvement of MAP compared with previous best) as well as the state-of-the-art on MNIST. Besides, we show that the hashing learned by CNN is better than KSH \cite{KSH} which took features from the layer prior to the selected fully connected layer from our model as input, and is also better than Euclidean-distance-based ranking with the features from the layer same with us as descriptor.

This paper is organized as follows: Section 2 briefly reviews previous related works of others. And the methodology of our work is revealed in Section 3. The experiments and discussions are presented in Section 4 and 5. Finally, we conclude the whole paper in Section 6.

\section{Related Work}
In the following subsections, we will briefly review hashing and CNN over the literature.

\subsection{Hashing Methods}
Recently, as the ever-growing web data makes information retrieval and other problems more challenging, hashing has become a popular solution \cite{CNNH} \cite{SPLH} . The short binary codes generated by hashing make retrieval efficient both on storage and computation. In many cases, search in millions of data will only consume constant time via tens-of-bit representations mapped from the query by hashing.

To generate $n$-bit code, hashing methods need $n$ hash functions the $k_{th}$ of which generally takes the following form:
\[
h_k(x) =\begin{cases}
1 &f_k(x)\geq b_k \\
0 & f_k(x)<b_k
\end{cases}
\]
where $x$ is a data sample, $f_k$ is the projection function, and $b_k$ the corresponding threshold. Based on the method to get $f_k$, hashing can be divided into data-independent and learning-based. And learning-based hashing methods can be classified according to using label information or not into unsupervised and supervised methods.

\subsubsection{Data-independent Hashing}
In the early stage of hashing, methods were mostly data-independent. For example, a typical category of Locality Sensitive Hashing (LSH) \cite{SSHDH} uses random projection to construct hash functions. The property of LSH, that samples within short Hamming distance in hash space are more likely to be near in their source space, makes it very attractive. But the metrics are asymptotically preserved with increasing code length. Thus, to achieve high precision, LSH-related methods require large hash tables. And LSH works only with theoretic guarantees for some metric spaces, such as Euclidean distance.

\subsubsection{Learning-based Hashing}
Unlike data-independent hashing, learning-based methods attempt to capture the inherent distribution of the task domain by learning. 

Unsupervised methods use only unlabeled data as training set, among which  are methods such as Kernelized LSH \cite{KLSH}, Semantic Hashing \cite{DAuto}\cite{SemH}and Spectral Hashing \cite{SH}. Semantic Hashing uses stacked RBMs (Restricted Boltzmann Machines) to learn binary hash codes from raw inputs. After pretrained layer by layer, the RBMs are unrolled and refined as a deep autoencoder. Spectral Hashing defines similarity on the feature space and attempts to minimize the weighted average Hamming distance between similar samples.

Supervised hashing utilizes more direct similarities such as human-annotated labels to get satisfying codes. KSH is a supervised method which uses kernel-based model to minimize the Hamming distances of learned hash codes between similar data samples while maximize the distances between dissimilar samples. BRE (Binary Reconstruction Embedding) \cite{BRE} learns hash functions by minimizing the reconstruction error between original distances and the Hamming distances of the corresponding binary codes. And BRE has its kernelized version. While initially proposed as unsupervised hashing, BRE is easy to extend as a supervised one by setting similar pairs with zero distance and dissimilar pairs one.

As a method of semi-supervised hashing, SPLH (Sequential Projection Learning Hashing) \cite{SPLH} can leverage unlabeled data as well as the relations between pairs of samples annotated with "similar" and "dissimilar" to iteratively get good hash functions which will compensate for the violation brought about by previous functions.

Except Semantic Hashing, these methods are kind of shallow and usually leverage some feature extraction algorithm to get the descriptors needed as inputs. But the relationships between samples in semantic space are not maintained by low-level representations. And even combined with high-level features, the conventional hashing methods are very likely to perform no better than a end-to-end system which learns the feature extractor and hash functions together. As for Semantic Hashing, even stacked into a deep structure, the thorough unsupervised training procedure makes it inferior in such an evaluation system like image retrieval.

\subsection{Convolutional Neural Network}
Since 2012, explosive interests in computer vision have been attracted by CNN \cite{Imagenet}. Its remarkable successes in kinds of tasks such as object recognition \cite{Imagenet}\cite{DSN}\cite{VGG}, detection \cite{Imagenet}\cite{VGG}, image parsing \cite{YL-Scene} and video classification \cite{VClaCNN} have push the gap between machine and human vision narrow down by a large step. 

CNN \cite{CNN} is a constrained multilayer neuron network whose inputs lie on 2-dimensional planes. Inspired by human visual system, the neurons in CNN's hidden layers (except the final classification part) take inputs from a local region of the previous layer and are tiled in 2d feature maps with respect to their input region. A typical CNN is constructed with three kinds of layers, i.e. convolutional layer, pooling layer and fully connected layer. And the layers are organized just like a stack whose lower parts are convolutional layer and pooling layer alternately and top layers are fully connected. Neurons within a convolutional feature map share weights with each other. Being placed behind convolutional layers, pooling layer can be divided into several classes according to the operation it takes, max-pooling and average-pooling for example. Both convolutional layer and pooling layer can be overlapped or not by adjusting the stride and input filter size.

Although filter weights of neurons within a same feature map are shared, the amount of parameters in a large CNN is still big enough to overfit the training data. Weight decay can only alleviate it in some cases. By randomly set a portion of activations of a layer with probability $p$ to zero at every training iteration and multiply the outgoing weights by $p$ during test, dropout \cite{Dropout} works as if thousands of models vote. Thus it can reduce overfitting greatly. Dropconnect \cite{DropConnect} is extended from dropout but consumes much more memory.

Another apparent way to overcome the overfitting problem is to increase the training data by annotating new samples or data augmentation. ImageNet \cite{imagenet_cvpr09} is a large image database with totally 14 million annotated images. With a deep CNN, Google has set a new record on LSVRC2014 classification track \cite{GoogleBatch}. 

It has been suggested that the features in deep layers learned from ImageNet possess great ability to represent visual content of images, and can be used for different tasks, such as scene parsing, detection and image retrieval \cite{ScenePVGG} \cite{RCNN} \cite{NeuralCodes}. Neural codes \cite{NeuralCodes} uses activations at a top layer from an ImageNet-pretrained CNN as descriptors of the corresponding input image. And then Euclidean distances are computed to measure similarities in semantic space. When retrained with datasets related to the query field, the retrieval performance can be comparable with state-of-the-art. Different from this work, we proved that the signs of these activations themselves were very informative. And we used a model fully trained with the given data and their labels, instead of pretrained by another dataset.

\subsection{Hashing with CNN}
Similar with \cite{DAuto}, CNNH/CNNH+ \cite{CNNH} take raw image data as input, but the latter two divide the learning process into two different stages. In the first stage, similarity/dissimilarity matrix $S$ is decomposed into a product $S=\frac{1}{q}HH^{T}$ where the $i$-th row in $H$  is the target binary codes for the $i$-th training image. In the second stage, the raw image pixels, pre-generated binary codes $H$ (CNNH+ together with their one-hot binary labels $Y$) are fed to a CNN whose objective is to minimize the error between outputs and the target binary vector concatenated by $H$ and $Y$.

But the decomposition stage would bring about extra errors. For example, the sum of squared errors (SSE) to reconstruct the 5000-by-5000 similarity matrix with 48-bit hash codes generated by gradient descent in \cite{CNNH} (the codes are obtrained from the author's web page) could be $1.8*10^7$, amount to $18\%$ of the possibly maximum SSE (e.g. $5000^2*2^2$).

On the other hand, it's not intuitive to generate approximate hash codes first and then train a CNN with these codes as target. In contrast, our proposed method took the semantic meaning as target.

\section{Methodology}
Normally, suppose that the $l$-th layer of a CNN is convolutional layer of stride 1, the convolution kernel between the $m$-th feature map in this layer and the $n$-th feature map in previous layer is $k^l_{m,n}$, the size of convolution kernel is $s$ and the amount of feature maps in $l$-th layer is $c_l$, then the response of a neuron on the $m$-th feature map in this layer at $(i,j)$ (counting from $(0,0)$)can be formulated as follow:
\begin{equation}
x^{l,m}_{i,j}=\sum\limits_{n=1}^{c_{l-1}}\sum\limits_{p=0}^{s}\sum\limits_{q=0}^{s}w^{l,m,n}_{p,q}\times x^{l-1,n}_{i+p,j+q}+b^{l,m}
\end{equation}
where $w^{l,m,n}_{p,q}$ is the weight at $(p,q)$ in convolutional layer kernel $k^l_{m,n}$, and $b^{l,m}$ is the shared bias for the $m$-th feature map in $l$-th layer \cite{NotesCNN}. Both $w^{l,m,n}_{p,q}$ and $b^{l,m}$ are learnable. If no padding is added to the input of the $l$-th layer, the side length of output feature maps will be smaller by $s-1$ than input. From the above formulation we can see that weights are shared within a feature map in $(l-1)$-th layer and just one bias is assigned to each feature map in $l$-th layer. Thus the number of parameters in a convolutional layer is $s*s*c^{l-1}*c^{l}+c^l$.

Operations in convolutional layer are linear, being multiplication and summation. But with algebra we can know that a multilayer network constructed with linear operations alone can be reduced to a one-layer operation. Nonlinear transformations are needed to endow networks the capacity to be deep. ReLU (Rectifier linear unit, $x=max(0,x)$), $tanh$ and $sigmoid$ are three mainly-used nonlinear transformations. ReLU is unsaturated and thus doesn't bring about the vanishing gradient problem, so can speed up training. In addition, ReLU is fast to compute.

Similar to convolutional layer, all the neurons in pooling layer take corresponding small patches from feature maps in previous layer as input. For example, we can formulate a non-overlapping max-pooling layer whose pooling size is $s$ as follows:

\begin{equation}
x^{l,m}_{i,j}=\max\limits_{p=0}^{s}\max\limits_{q=0}^{s}x^{l-1,m}_{i\times s+p,j\times x+q}
\end{equation}
where parameters are defined as for convolutional layer. If the stride equals to $s$	, the size of feature maps will be decreased by a factor of $s$ and the number of feature maps keeps unchanged. Pooling layer has no parameters and the pooling stride is often larger than 1, so the speed of feed-forward and back propagation within pooling layer is fast and can reduce the downstream calculation. Max pooling chooses the maximum value in input patch as output, while average pooling takes the average response as output.

A classifier is then connected to the stack of convolutional and pooling layers. Usually, logistic regression with negative log likelihood as loss function is used for single-label classification. Other kinds of loss functions such as Euclidean distance and cross-entropy are also in common use.

Then during training, classification errors are back propagated to update the learnable parameters. Methods such as weight decay and dropout can be used to alleviate the overfitting problem which is caused by that a relatively large CNN is used to train on a small dataset. Moment can help with oscillation.

\subsubsection{The model}As we put it in the Introduction, the CNNH/CNNH+ can be considered as a Siamese structure cut at the final layer, which is an inner product operation. So it is still a conventional classification system, except that it is in a fashion of stack and doesn't refine the system as a whole. So we consider its performance may be worse than a holonomic Siamese structure. But the amount of possible image pairs labeled with "similar" or "disimilar" needed by Siamese structure is quadratic of the images' number, and thus expensive to annotate and train. We assumed that the similarity/dissimilarity matrix constructed with two discrete values that indicate similar or dissimilar is not necessarily more informative than a class matrix. So we got a rather straightforward way to train a model for hashing, i.e. the classic CNN targeted at classification.

\subsubsection{Hash codes}For any neuron in a CNN, the activation is a notification of some unknown patterns. \cite{RCNN}\cite{visualization_Bengio} look through the dataset for images that maximize the activation of the neuron, taking the neurons as meaningful filters. This paper adopted this view. And we hypothesized the signs of activations of neurons were able to represent whether there is a certain image pattern at the corresponding receptive field. Because binary outputs will not back propagate gradients to the prior layers, we didn't take any modification into the CNN model for this in training. The hash codes were binarilized from the activations of chosen layer with threshold of zero, i.e. taking the below expression as the ultimate hash codes:
\[
h_k(x) =\begin{cases}
1 & x^{l,k}\geq 0 \\
0 & x^{l,k}<0
\end{cases}
\]
where $x$ is the input image and $x^{l,k}$ the maximum activation of $m$-th feature map in chosen $l$-th layer (if the chosen layer is fully connected, each neuron is considered as a feature map). While geographical information does help with classification and other vision tasks, we were just going to discard these information by globally max-pooling or utilizing a fully-connected layer's outputs as hash codes.

\subsubsection{Chosen layer}Previous works \cite{DeConv}\cite{transferable} have shown that features learned in the shallow layers of a CNN are low-level and general, while those in the deep layers are more abstract and related to particular classes. Either too general or too abstract, the information entropy of the layer will be far from optimal. Taking the CNN's final output layer as an example, suppose that the training data are evenly distributed among $n$ classes, then the output bit of one class is only activated with a possibility of $\frac{1}{n}$. Thus the information contained is only $\frac{1}{n}log n + \frac{n-1}{n}log \frac{n}{n-1}$ bit. The bigger $n$ is, the less information a bit contained. Features too general are even worse. So we tried with middle layers of CNN for hashing evaluation.

Consider CNN as a directed acyclic graph, then we suppose that the selected layers should better be a cut of the graph so as to possess sufficient representative power. 

As it is not easy to train a single-path model for features from global max-pooling, we trained a model with multi path to evaluate the performance of middle convolutional layer (similar to the model used in our CIFAR-10 experiment). Outputs of the selected middle convolutional layer were fed into a global max-pooling layer and bypassed the following layers to concatenate with outputs of the final pooling layer, then the resulted concatenation was connected with the first fully connected layer. But the performances of the binary codes generated from this kind of layers were very poor (the mean average precision of approximately 0.2) and it was not so convenient to control as the fully connected layers, we will not report it in the experiments part.
	 
Then, we selected the first fully connected layer. To control the length of hash codes, another fully connected layer with no nonlinear transformation was inserted between. Let the number of newly joined units be $m$. and the number of subsequent layer's neurons $n$. It can be proved that when $m$ is no less than $n$, the two layers were equivalent with the original one. Because each bit in the hash code should be uncorrelated with others, we put a dropout layer between the two layers. The drop probability $p$ was a hyper parameter which was related with dataset and the code length. Intuitively, as long as $m*(1-p)$ is still larger than $n$, the performance will not be degraded.

\section{Experiments}
Our models were implemented with cuda-convnet2\footnote{https://code.google.com/p/cuda-convnet2/}, an open-source CNN tool developed by Alex Krizhevsky. The description of an architecture is given in the following way :1x28x28-32C5P2-MP3S2-32C5P0-H32-D0.5-H10 represents a CNN with inputs of 1 channel of $28\times 28$ pixels, a convolutional layer with 32 filters whose size is $5\times5$ and 2 paddings around the input maps, a max pooling layer of $3\times3$ size and stride 2, a 32-filters convolutional layer whose kernel size is $5\times5$ and have 0 padding around the input maps, a hidden layer with 32 neurons, a dropout layer whose possibility to dropout is 0.5, and a hidden layer of 10 neurons, finally a softmax layer.

We denoted our proposed hashing method as CNNBH, and compared it with LSH, BRE, KSH, CNNH and CNNH+ on two datasets, i.e., MNIST \footnote{http://yann.lecun.com/exdb/mnist/} and CIFAR-10\footnote{http://www.cs.toronto.edu/~kriz/cifar.html} \cite{80mTiny}, both of which are widely used in the literature of hashing for image data.

The MNIST dataset of hand-written digits consists of 70,000 gray images. Each sample is of size $28\times 28$ pixels and annotated with one label from 0 to 9. And the CIFAR-10 dataset is a subset of the 80-million tiny images collection. Containing 60,000 color images of $32\times 32$ pixels, the CIFAR-10 is evenly labeled with 10 mutually exclusive classes, ranging from airplane to bird. 

To compare on an equivalent basis, we sampled from each dataset 1000 images as query set and another 5000 from the rest for training like \cite{CNNH} \footnote{Different from CIFAR-10, we didn't get the index of the 5000 training data on MNIST, so a different sample set was used. But according to Pan Yan, the results of CNNH are relatively insensitive to data partitioning.}. For LSH and KSH, all the data except the query set were training set. The 5000 keeping-label samples served as training set for the other methods. Note when tested on MNIST, the raw data were input to all approaches, but with CIFAR-10 the raw images and 512-dim GIST features were inputed to CNN based methods and the others respectively.

As for our proposed CNNBH, the 5000 labeled samples in each datasets were divided into 5 folds to do cross-validation to find the most suitable models which were then trained with all the 5000 samples. By "suitable" we mean that the model's classification precision was good as well as the complexities were comparable with Gist feature extraction algorithms.

Hash lookup and Hamming ranking are two widely used method to conduct search with hashing. Hash lookup constructs a lookup table with radius $r$ in advance, and all the samples fallen within the radius will be returned as results, thus can decrease the query time to a constant value. However, the number of results returned will dramatically decrease with the code's length increases. On the other hand, Hamming ranking will traverse the dataset all through at a new coming. 

We evaluated the performances on three metrics, i.e. Precision curves within Hamming radius 2, Precision-Recall curves and Precision curves with respect to different returned number with ranking.

For LSH, the projections were randomly sampled from a Gaussian distribution with zero-mean and identity covariance to construct the hash tables. As for BRE and CNNH+, we used the best setting reported in the literature \cite{CNNH}.

\subsection{MNIST}
Because the MNIST is less challenging, we chose a two-conv-layer structure to generate the binary codes. To get 32-bit codes, a 1x28x28-32C5P0-MP2S2-32C5P0-MP2S2-H32-D0.5-H10 model is used. With respect to other code length, we just need to change the "H32" layer with the corresponding number of neurons, and tune the drop probability of ¡°dropout¡± layer. We should call "H32" layer the knob.

In our experiments, the drop probability of the final dropout layer was not a sensitive factor. So we empirically set the probabilities to dropout in all the seven models to 0.5. Rectifier Linear Unit (ReLU) was used as the nonlinear transformation neurons for the two convolutional layers. Before training we randomly initialized the weights of each layer with Gaussian distributions whose $\mu$ were all 0 and $\sigma$ 0.2, 0.2, 0.01 and 0.1 for layers from shallow to deep respectively. During training, the learning rate was firstly set to 0.01 for 50 epochs and then multiplied by one tenth after every 20 epochs. We trained the networks for 90 epochs in total.

The MAP of Hamming ranking can be seen on Table 1 and the performance curves are shown in Figure 3. Figure 3(a) shows the precision of returns with hash lookup of radius 2. From this figure we can know that along with the code length increasing, in contrast with other models, our models performed better and better. The LSH reported in this paper used random projection which relies on the input feature to maintain the similarity. But because of the semantic gap between low-level feature and content, LSH's performances were not stable. While BRE and KSH performed better than LSH, they were still worse than CNN based methods e.g. CNNBH, CNNH and CNNH+. Figure 3(b) and Figure 3(c) show the performances of 48-bit codes generated by all five models.

On this dataset, our models performed barely inferior to CNNH and CNNH+. But given that our models were much smaller with just two convolutional layers and no local response normalization layers, and that the MNIST dataset is a rather easy one, we took it a straightforward contrast.

\begin{figure*}[ft]
\centering
\subfigure[]{
\includegraphics[width=0.3\textwidth]{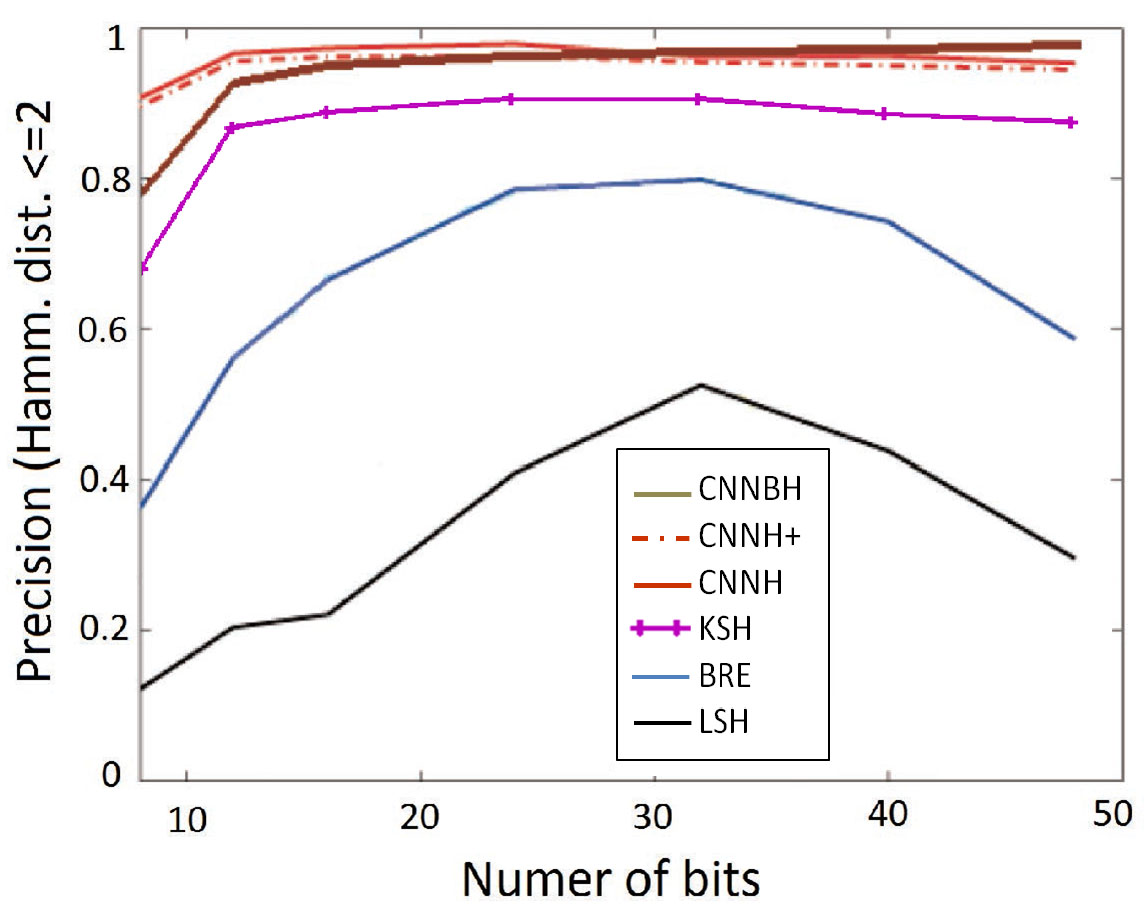}
}
\subfigure[]{
\includegraphics[width=0.3\textwidth]{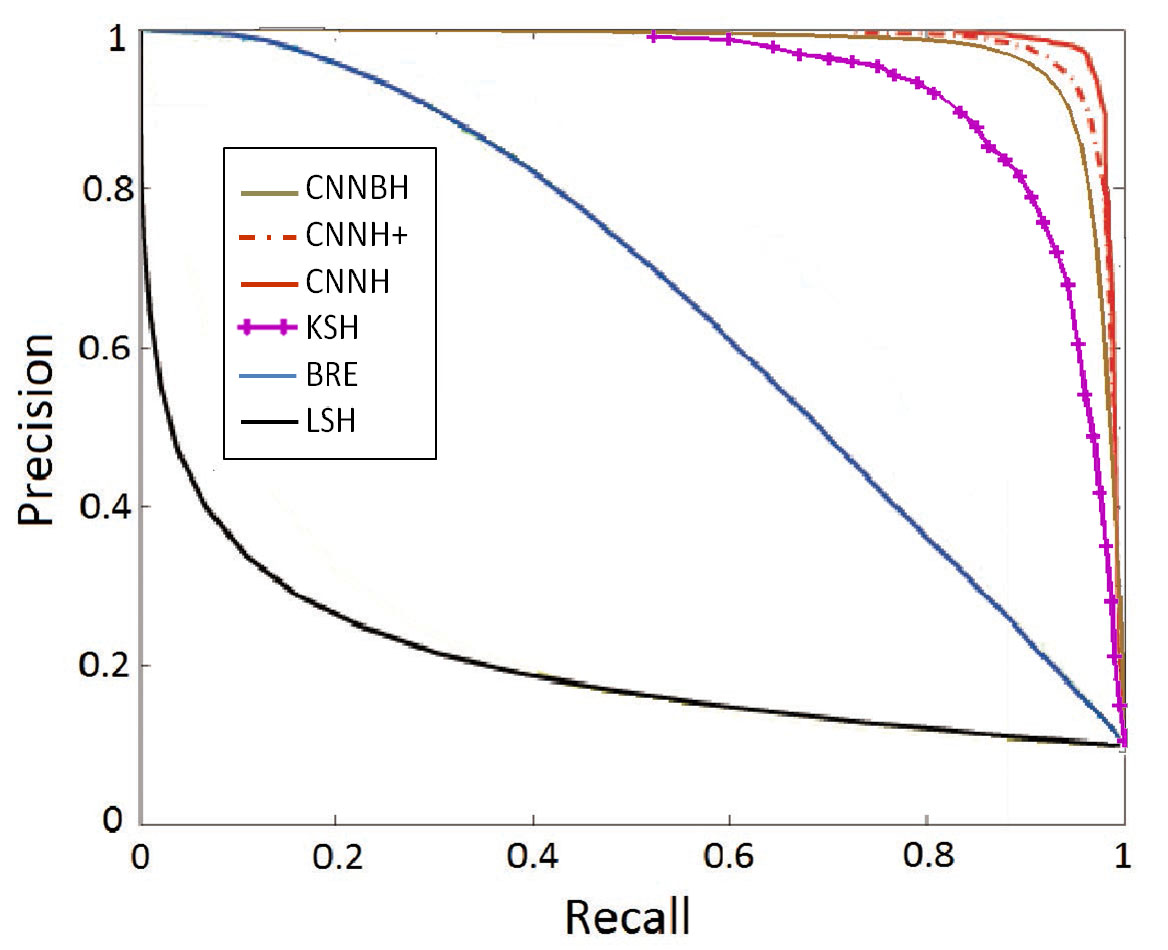}
}
\subfigure[]{
\includegraphics[width=0.32\textwidth]{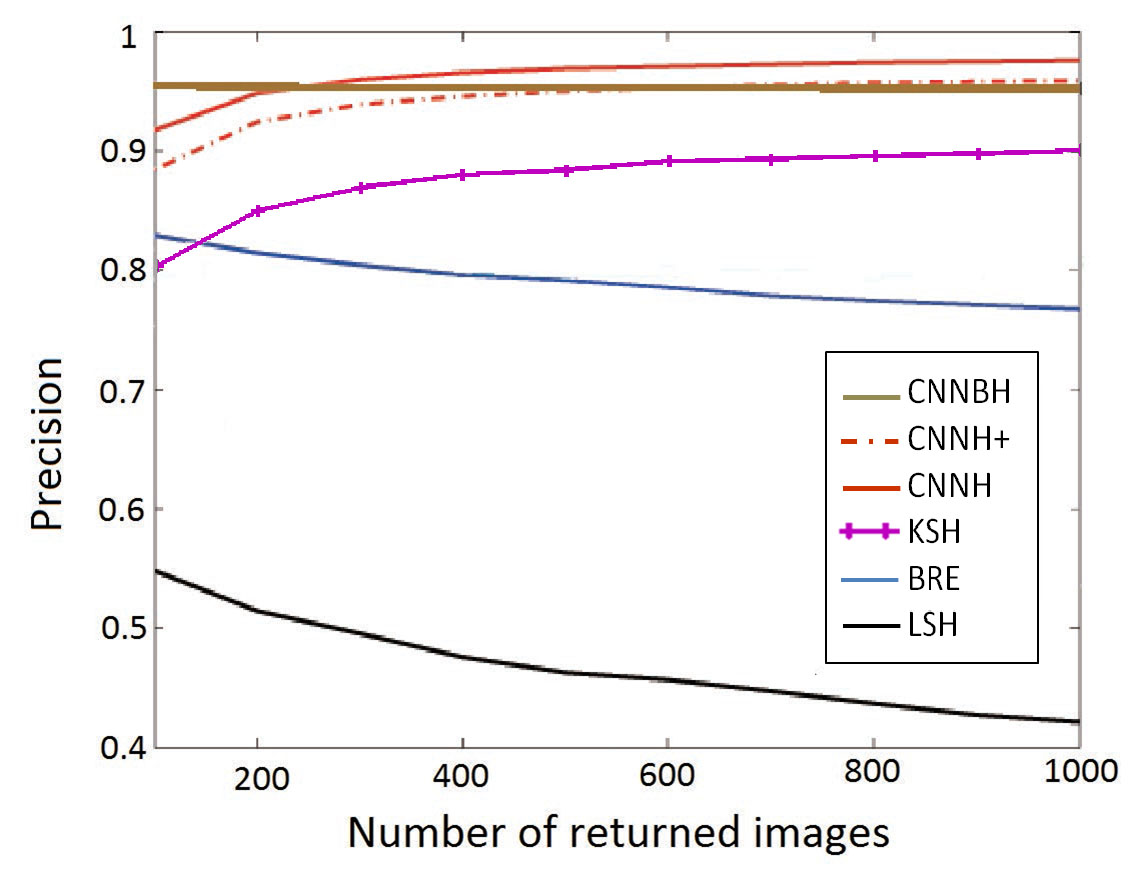}
}
\caption{The results on MNIST. (a) precision curves of hash lookup within Hamming radius 2. (b) precision-recall curves of Hamming raking with code's length of 48 bits. (c) precision curves with respect to number of returned images of Hamming raking with code's length of 48 bits. (Adapted from \cite{CNNH})}
\end{figure*}

\begin{figure*}[t]
\centering
\subfigure[]{
\label{fig:subfig:a}
\includegraphics[width=0.28\textwidth]{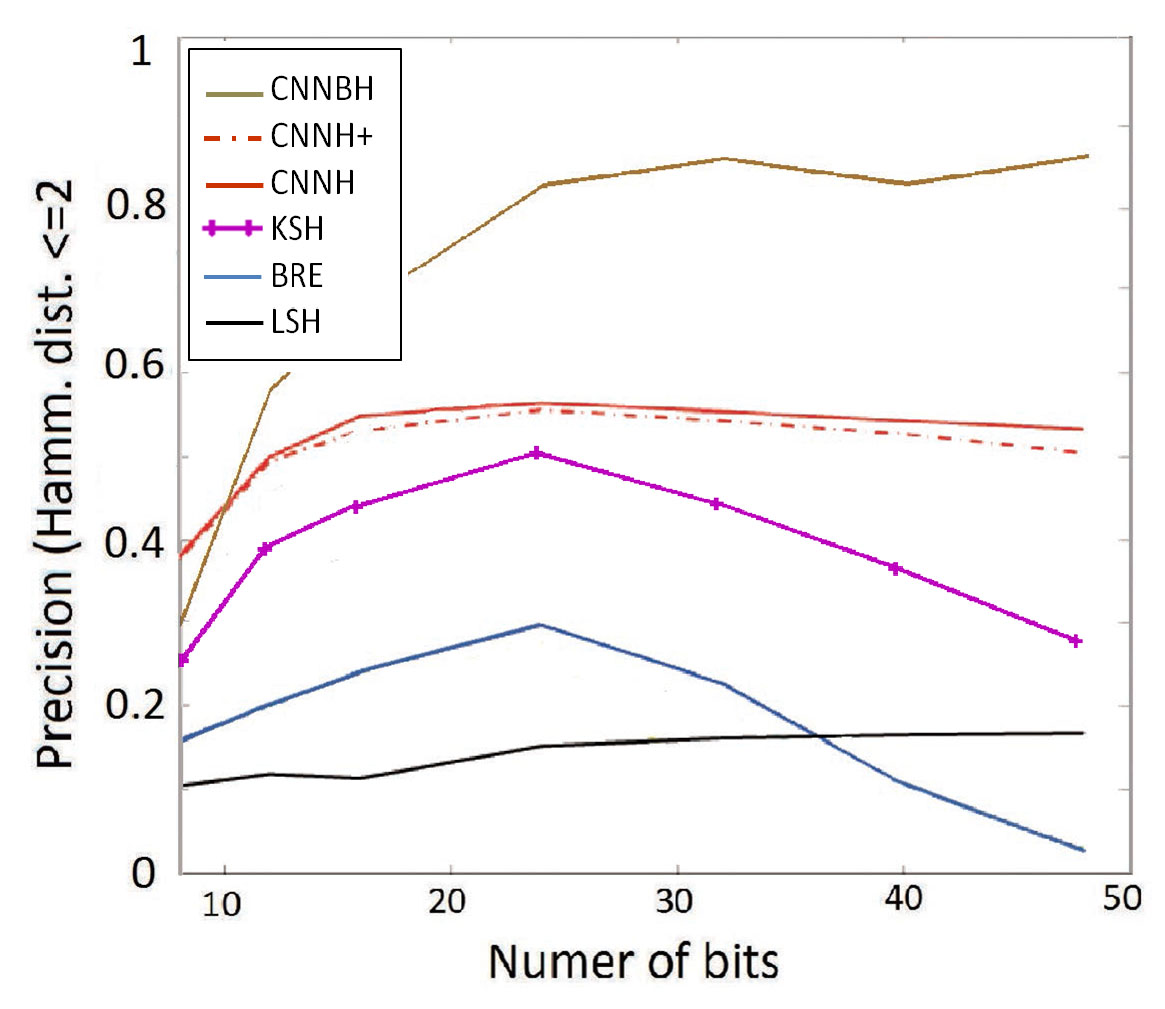}
}
\subfigure[]{
\includegraphics[width=0.3\textwidth]{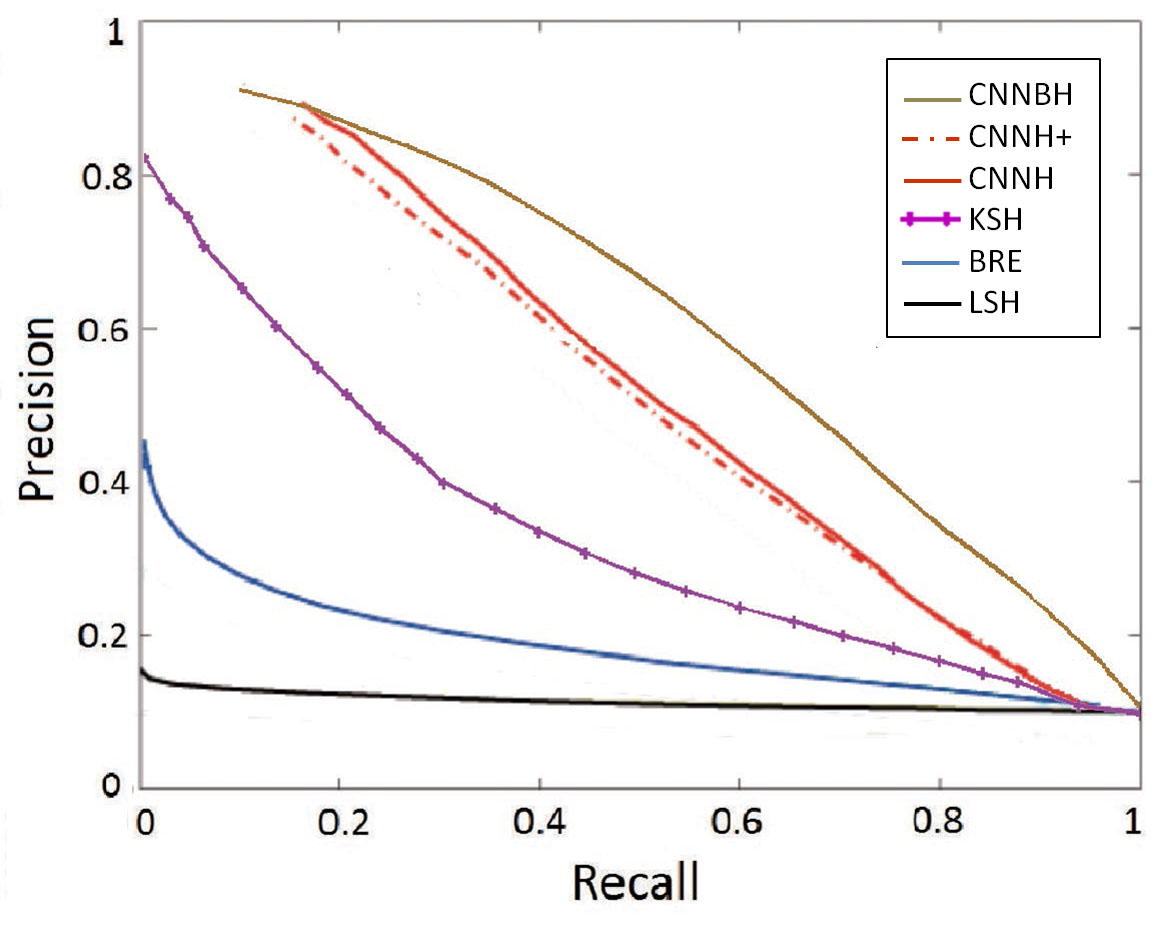}
}
\subfigure[]{
\includegraphics[width=0.3\textwidth]{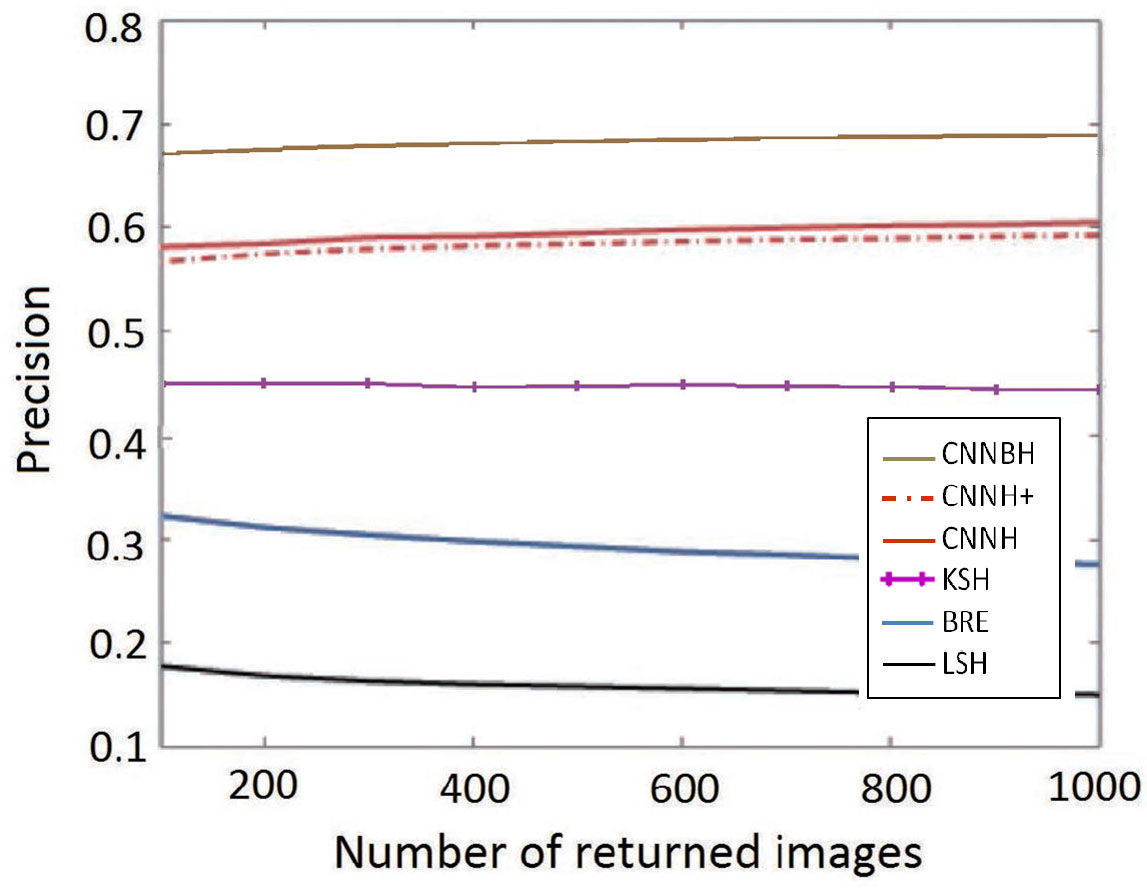}
}
\caption{The results on CIFAR-10. (a) precision curves of hash lookup within Hamming radius 2. (b) precision-recall curves of Hamming raking with code's length of 48 bits. (c) precision curves with respect to number of returned images of Hamming raking with code's length of 48 bits. (Adapted from \cite{CNNH})}
\end{figure*}

\begin{table}
\centering
 \renewcommand{\arraystretch}{1.1}
\caption{MAP of Hamming ranking w.r.t different number of bits on three datasets.}
\begin{tabu}{X[1.5C]|X[C]X[C]X[C]X[C]|X[C]X[C]X[C]X[C]}
\hline
Method & \multicolumn{4}{c}{MNIST(MAP)} & \multicolumn{4}{|c}{CIFAR10(MAP)} \\
 code length& 12bits & 24bits & 32bits & 48bits & 12bits & 24bits & 32bits & 48bits \\
\hline
CNNBH & $0.946$ & $0.952$ & $0.953$ & $0.939$ & $\bm{0.532}$ & $\bm{0.564}$ & $\bm{0.610}$ & $\bm{0.617}$ \\
\hline
CNNH+ & $\bm{0.969}$ & $\bm{0.975}$ & $\bm{0.971}$ & $\bm{0.975}$ & $0.465$ & $0.521$ & $0.521$ & $0.532$ \\
\hline
CNNH  & $0.957$ & $0.963$ & $0.956$ & $0.960$ & $0.439$ & $0.511$ & $0.509$ & $0.522$ \\
\hline
 KSH  & $0.872$ & $0.891$ & $0.897$ & $0.900$ & $0.303$ & $0.337$ & $0.346$ & $0.356$ \\
\hline
 BRE  & $0.515$ & $0.593$ & $0.613$ & $0.634$ & $0.159$ & $0.181$ & $0.193$ & $0.196$ \\
\hline
 LSH  & $0.187$ & $0.209$ & $0.235$ & $0.243$ & $0.121$ & $0.126$ & $0.120$ & $0.120$ \\
\hline
\end{tabu}
\end{table}

\subsection{CIFAR-10}
We used a structure larger than the MNIST model but with smaller filter sizes at this dataset to generate 32-bit length codes: 3x32x32-32C3P1-32C1P0-MP3S2-D0.5-32C3P1-32C3P1-MP3S2-D0.5-64C3P1-64C3P1-MP3S2-D0.5-H32-D0.5-H10. For the convenience of later quotation, we named the third pooling layer "Pool3" and the first fully connected layer "H32". Similar to MNIST structure, "H32" layer was the knob. In this experiment, ReLU was utilized in all the six convolutional layers as nonlinear units. Weights of both convolutional and fully connected layers were initialized randomly from a Gaussian distribution of 0 $\mu$ and 0.04 $\sigma $.  But the probability of the final dropout layer to drop was important. When the number of neurons in the knob layer decreased smaller than 20, possibility of 0.5 in the final dropout layer enlarged the error rate on validation set. To overcome this, we reduced the drop probability bit by bit until the error rates were similar to that of the 48-bit model. After trials, we got the pairs of the probability and code length as Table 2. Because CIFAR-10 contains much more complicated background and transformations than MNIST and we supposed 5000 was far from saturation, we undertook more epochs to train the models of CIFAR-10 sufficiently. Firstly we trained the models with learning rate of 0.01 for 500 epochs, and then decreased the rate to be one tenth of by every 200 epochs. 900 epochs were executed for all seven models.

Despite that the precision of models were inferior to CNNH's and CNNH+'s when encoded with less bits, Figure 4(a) witnessed that precision of images within 2 bits Hamming distance returned by our models increased along with the length of hash codes. And in Figure 4(b) the precision-recall curve of our model enclosed more spaces than the others, which suggested that the performance of our CNNBH was the best. Actually, the mean average precision (MAP) of our 48-bit model was 0.617 in comparison with the previous record 0.532 created by CNNH+. Then Figure 4(c) demonstrated our model's best performance on Hamming ranking that our model improved the precision from about $60\%$ to $69\%$, nearly by $15\%$ beyond CNNH+. And again, we should mention that the CNN architecture of our models contained much less parameters than CNNH+, which was 125K in comparison with 318K.

\begin{table}[t]
\centering
\renewcommand{\arraystretch}{1.1}
\caption{Pairs of code length and drop probability of the final dropout layer in CIFAR-10 model}
\begin{tabu}{X[2.4]|X[C]|X[C]|X[C]|X[C]|X[C]|X[C]|X[C]}
\hline
{code length } & 8 & 12 & 16 & 24 & 32 & 40 & 48\\
\hline
{dropout probability}& 0 & 0.2 & 0.3 & 0.5 & 0.5 & 0.5 & 0.5\\
\hline
\end{tabu}\vspace{.3cm}
\end{table}

\begin{table}[t]
\centering	
\renewcommand{\arraystretch}{1.3}
\caption{Ranking precisions of different methods at 500 and 1500}
\begin{tabu}{X[0.63l]|X[1c]|X[1c]}
\hline
\diagbox{Method}{Precision}{Top N} & 500 & 1500 \\
\hline
CNNBH & $\bm{68.17\%}$ & $\bm{69.06\%}$ \\
\hline
L2\_fc & $61.50\%$ & $58.88\%$ \\
\hline
L2\_fc+ & $62.62\%$ & $60.51\%$ \\
\hline
Pool3\_feat + KSH & $67.02\%$ & $67.88\%$ \\
\hline
Pool3\_feat + LSH & $39.33\%$ & $34.64\%$ \\
\hline
Gist\_320 + LSH & $20.60\%$ & $18.77\%$ \\
\hline
\end{tabu}
\end{table}

\subsection{Anylasis on CIFAR-10}
On CIFAR-10, we conducted some other experiments to compare. One was image retrieval using Euclidean distance between corresponding features of query and dataset's images as similarity, and the features were right the outputs of "H32". To show that our model performed best not just because the features extracted by CNN were more powerful, another experiment we undertook was KSH which performed best on Gist features with outputs of "Pool3" as descriptors of an input image. The results are shown in Table 3, in which L2\_fc and L2\_fc+ were both Euclidean distance, but the features used by L2\_fc+ were rectified as follow:

\[
X_{fc+} =\begin{cases}
X_{fc}+0.5 & X_{fc}\geq0 \\
X_{fc}-0.5 & X_{fc}<0
\end{cases}
\]
where $X_{fc}$ represents the original outputs of "H32".
Surprisingly, we found both L2\_fc and L2\_fc+ performed next to CNNBH, and that the performance of L2\_fc+ was slightly higher than L2\_fc. This may suggest that the differences within feature values of same sign are not as informative as the sign itself. KSH learned hash functions from features extracted from "Pool3", right the preceding layer of the knob. The retrieval precision of "Pool3\_feat + KSH" was inferior than our model by a small margin.in comparison with the other three methods demonstrated that our model performed good not just because the features learn by CNN are powerful. And the last two were random-projectioin-based LSH with the "Pool3" feature and 320-dimension Gist as input respectively. From the comparison between the two LSH-based hashing methods, we can infer that the features learned by CNN is indeed better than Gist. And the inferior performance of "Pool3\_feat + KSH" demonstrated that our model works not just because of the CNN-features, because KSH used kernelized method while our model can be seen as concatenating a linear projection to "Pool3".


\section{Discussion}
Figure 3(a) and Figure 4(a) showed that the precision of the proposed CNNBH on both MNIST and CIFAR-10 within Hamming radius of 2 improved with more hash bits, which is consistent with human intuition.

In the case of a space where visually similar images are distributed messily, one linear projection is hardly able to reduce the entropy of data samples. In fact, when samples are completely randomly distributed, any number of projections is unable to reduce the entropy and thus can't achieve good retrieval performance. Along with data organizes structured, the performance of learning-based projection improves. LSH can get persistent good performance only when similarity relationship is kept in the original space.

The chosen fully connected layer can be considered as learning-based linear projections, whose inputs are also learnable. During training, the weights and biases in CNN are adjusted to render the feature maps in deep layer expressive enough so as to allow the simple classifier behind perform well. And the raw pixels are at the same time transformed into a description space where similarity relationships are correlated with Euclidean distance.

Except that the features in our model are more powerful, we think the decomposition in CNNH/CNNH+ would bring about extra errors which can make the training target diverge from minimizing the error with related to the similarity/dissimilarity matrix. In contrast, our method takes classification as the optimization objective in which the information contained is equivalent to S. And the optimization objective is then transferred by back propagation into learning expressive filters.

And, compared with CNNH which is constructed with two separate stages, our model is trained end-to-end. We suppose it is the co-adaption between layers that results into the performance improvement on CIFAR-10. KSH with CNN-feature performs closely to our method, but it is kernelized and thus slower than our model.

\section{Conclusion}
As a method of ANN search, interests of many researchers as well as companies have been attracted to hashing. We proposed a new method which can obtain the binary hash code of a given image just by binarilizing outputs of a certain fully connected layer (in our experiments, we chose the first fully connected layer), and achieved the best result on CIFAR-10 as well as the state-of-the-art performance on MNIST.

Unlike CNNH and CNNH+ which need matrix decomposition, our method trained the CNN from raw pixels in a normal process. So more training data will not burden it too much. In consideration of the fact that all the models use only 5000 samples to train, we can expect an improvement of performance with a larger training set. Besides, a large amount of unlabeled data remains untouched, which can be another key element to enhancement. And we consider our CNNBH and CNNH are both simpification of Siamese structure, so it is another field to dwell on.

\bibliographystyle{abbrv}
\bibliography{sigproc} 



\end{document}